\documentclass[final,5p,twocolumn]{elsarticle}
\usepackage{xcolor} % or package xcolor
\usepackage{booktabs} 
\usepackage{float}
\usepackage{caption} 
\usepackage{subcaption} 
\usepackage[utf8]{inputenc} 
\usepackage{url}
\usepackage{listings}
\usepackage[frozencache=true,cachedir=minted-cache]{minted}% \usepackage{verbatimbox}
\usepackage{pifont}% http://ctan.org/pkg/pifont

\definecolor{codegreen}{rgb}{0,0.6,0}
\definecolor{codegray}{rgb}{0.5,0.5,0.5}
\definecolor{codepurple}{rgb}{0.58,0,0.82}

\usepackage{ascii}

\lstdefinestyle{mystyle}{
    language=Python,
    commentstyle=\color{codegray},
    keywordstyle=\color{magenta},
    numberstyle=\tiny\color{codegreen},
    stringstyle=\color{codepurple}\scriptsize\asciifamily,
    basicstyle=\scriptsize\asciifamily,
    breaklines=true,                 
    tabsize=2,
    morekeywords={True, False},
    basewidth={.55em}
}

\restylefloat{table}

\newcommand{\cmark}{\textcolor{green}{\ding{51}}}%
\newcommand{\xmark}{\textcolor{red}{\ding{55}}}%

\usepackage{array}
\newcolumntype{P}[1]{>{\centering\arraybackslash}p{#1}}

\usepackage[colorlinks = true, allcolors = blue]{hyperref} % \href

\journal{SoftwareX}

\begin{document}
\begin{frontmatter}



\title{tsflex: flexible time series processing \& feature extraction}


\author{Jonas Van Der Donckt \corref{contrib}}%\ead{jonvdrdo.vanderdonckt@ugent.be}
\author{Jeroen Van Der Donckt \corref{contrib}}%\ead{jeroen.vanderdonckt@ugent.be}
\author{Emiel Deprost}
\author{Sofie Van Hoecke}

\cortext[contrib]{Contributed equally}

\address{IDLab, Ghent University - imec, Technologiepark Zwijnaarde 126, 9052 Zwijnaarde, Belgium}

\begin{abstract}
%Ca. 100 words
Time series processing and feature extraction are crucial and time-intensive steps in conventional machine learning pipelines. Existing packages are limited in their applicability, as they cannot cope with irregularly-sampled or asynchronous data and make strong assumptions about the data format. Moreover, these packages do not focus on execution speed and memory efficiency, resulting in considerable overhead. % Add some more limitation that better transition to rest -> ook performance is niet topf
We present \texttt{tsflex}, a Python toolkit for time series \textbf{processing and feature extraction}, that focuses on performance and flexibility, enabling broad applicability. %ymultiple application %This is THE universal preprocessing and feature extraction toolkit you need when it comes to sequence data wrangling. :D //While making minimal assumptions about the series data, which greatly improves its usability. %that is flexible and domain-agnostic.
% This toolkit leverages sequence based arguments for strided-window feature extraction, and the sequence-index is maintained through all operations.
This toolkit leverages window-stride arguments of the same data type as the sequence-index, and maintains the sequence-index through all operations.
\texttt{tsflex} is \textbf{flexible} as it supports (1) multivariate time series, (2) multiple window-stride configurations, and (3) integrates with processing and feature functions from other packages, while (4) making no assumptions about the data sampling regularity, series alignment, and data type. Other functionalities include multiprocessing, detailed execution logging, chunking sequences, and serialization. Benchmarks show that \texttt{tsflex} is \textbf{faster} and more \textbf{memory-efficient} compared to similar packages, while being more permissive and flexible in its utilization.

\end{abstract}

\begin{keyword}
time series \sep processing \sep feature extraction \sep machine learning \sep data wrangling \sep python 


\end{keyword} 

\end{frontmatter}

\section*{Required Metadata}
\label{seq:metadata}
\noindent
Code: \href{https://github.com/predict-idlab/tsflex}{\texttt{github.com/predict-idlab/tsflex}} \\
Documentation: \href{https://predict-idlab.github.io/tsflex}{\texttt{predict-idlab.github.io/tsflex}} \\
License: \texttt{MIT} \\
Code version paper: \href{https://github.com/predict-idlab/tsflex/releases/tag/v0.2.3}{\texttt{0.2.3}}

\section{Motivation and significance}
\label{sec1:motivation}
% Introduce the scientific background and the motivation for developing the software.
Data-driven modelling and forecasting of time series is a major topic of interest in academic research and industrial applications, being a key component in various domains such as climate modelling~\cite{pieters2021mirra}, patient monitoring~\cite{topol2019highmedicine},  industrial maintenance~\cite{cook2019anomalyIoT}, and decision-making in finance~\cite{taylor2008modellingfinancial}.

Two traditional steps in machine learning on time series are (pre)processing and feature extraction, often performed in this order.
Processing is concerned with cleaning or transforming the raw data, e.g., filtering noise, detrending, clipping outliers, and resampling. 
Feature extraction aims to extract a set of characteristics, i.e., the features, with the intention of constructing a relevant (lower-dimensional) representation of the data. 
Both steps are time-consuming and rather complex, yet they are crucial for a successful machine learning pipeline~\cite{domingos2012few}. 

% Explain why the software is important, and describe the exact (scientific) problem(s) it solves.
In many cases the time series measurements might not necessarily be observed at a regular rate or could be unsynchronized~\cite{yadav2018mining}. Moreover, the presence or absence of measurements and the varying sampling rate may carry information on its own~\cite{little2019statistical}.
Unfortunately, current Python time series packages such as \texttt{seglearn}~\cite{burns2018seglearn}, \texttt{tsfresh}~\cite{christ2018tsfresh}, \texttt{TSFEL}~\cite{barandas2020tsfel}, and \texttt{kats}~\cite{noauthor_kats_nodate} make strong assumptions about the sampling rate regularity and the alignment of modalities. 
% In addition to this, none of these packages (conveniently) support stride-window segmentation.
Furthermore, to the best of our knowledge, no library today supports multiple strided-window feature extraction, varying data types (e.g., handling categorical data), and chunking of (multiple) time series. 
These observations highlight the need for a flexible processing and feature extraction package. Therefore, we present \texttt{tsflex}, a package designed solely concerning these two steps, as it aims to get the fundamentals right. 
%\texttt{tsflex} enables flexible and efficient (pre)processing and construction of features for time series machine learning. 
\texttt{tsflex} offers, next to custom functions, seamless integration with other data science packages, e.g., processing or feature functions from libraries such as \verb|NumPy|~\cite{harris2020array}, \texttt{SciPy}~\cite{2020SciPy-NMeth}, \texttt{seglearn}~\cite{burns2018seglearn}, \texttt{tsfresh}~\cite{christ2018tsfresh}, and \texttt{TSFEL}~\cite{barandas2020tsfel}, or machine-learning toolkits like 
\texttt{scikit-learn}~\cite{sklearn_api}. 
% We believe that this integration enables the construction of problem-specific features, e.g., statistical, temporal, and spectral.

%Indicate in what way the software has contributed (or how it will contribute in the future) to the process of scientific discovery; if available, this is to be supported by citing a research paper using the software.
\texttt{tsflex} can be employed from prototyping machine learning pipelines to deploying real-world time series\nolinebreak\hspace{\fill}\linebreak
projects. Currently, we are amongst others using \texttt{tsflex} in real-time data pipelines for the \textit{mBrain} study~\cite{brouwer_not_nodate}. Here \texttt{tsflex} is used for processing and feature extraction of raw sensor data streams in which gaps, irregular sampling rates and large data chunks occur.

%Introduce related work in literature (cite or list algorithms used, other software etc.).
The remainder of this paper is as follows. In section~\ref{sec2:description} we elaborate on the software and its functionality. Next on, section~\ref{sec3:examples} provides an illustrative example. Section~\ref{sec4:impact} stresses the impact of \texttt{tsflex} by both positioning our toolkit among existing libraries and benchmarking these libraries against \texttt{tsflex}. Finally, we end with a conclusion in section~\ref{sec5:conclusion}

%We observe that \texttt{tsflex} outperforms these suitable alternatives in both runtime and peak memory-usage, whilst even complying to the assumptions that those libraries make.

% To the best of our knowledge, \texttt{tsflex} is the first toolkit using time based window \& stride parameters for feature extraction and data chunking, resulting in an increased flexibility and more user convenience.

%Provide a description of the experimental setting (how does the user use the software?).
% moved to section 2

\section{Software description}
\label{sec2:description}
%Describe the software in as much as is necessary to establish a vocabulary needed to explain its impact. 
\texttt{tsflex} is a Python package that leverages (under the hood) efficient \texttt{NumPy}~\cite{harris2020array} data operations on \texttt{pandas}~\cite{reback2020pandas} data for (pre)processing and extracting features from time series. 
We opted for pandas data (either\nolinebreak\hspace{\fill}\linebreak \texttt{pd.DataFrame} or \texttt{pd.Series}) since this is a convenient format for sequence data, and supports amongst others sequence indexing, integrated column names, and various data types. A direct result of complying with the available \texttt{pandas} \href{https://pandas.pydata.org/pandas-docs/stable/user_guide/basics.html#dtypes}{data types} is that \texttt{tsflex} allows performing operations on numerical, categorical, boolean, time based, and string-like data.

Users can install \texttt{tsflex} by using \href{https://pypi.org/project/tsflex/}{\textit{pip}}; \texttt{pip install tsflex}, or \href{https://anaconda.org/conda-forge/tsflex}{\textit{conda}}; \texttt{conda install -c conda-forge}\nolinebreak\hspace{\fill}\linebreak \texttt{tsflex}. Once installed, our \href{https://predict-idlab.github.io/tsflex/}{documentation} together with various \href{https://github.com/predict-idlab/tsflex/tree/main/examples}{examples} should enable the user to apply this\nolinebreak\hspace{\fill}\linebreak toolkit for their purpose.

\subsection{Software Architecture}
\label{subsec:software_architecture}
\begin{figure}
\centering
\begin{subfigure}[b]{0.4\textwidth}
   \includegraphics[width=1\linewidth]{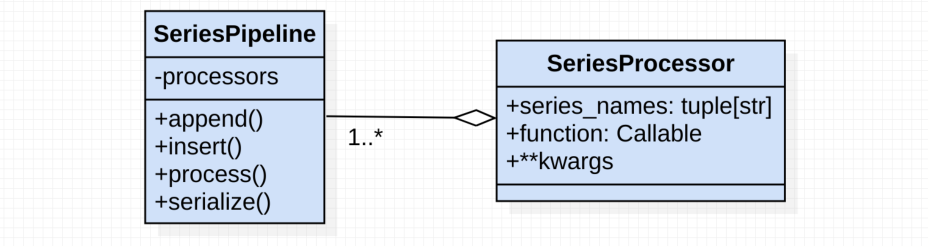}
   \caption{}
   \label{fig:processing_uml} 
\end{subfigure}

\begin{subfigure}[b]{0.5\textwidth}
   \includegraphics[width=1\linewidth]{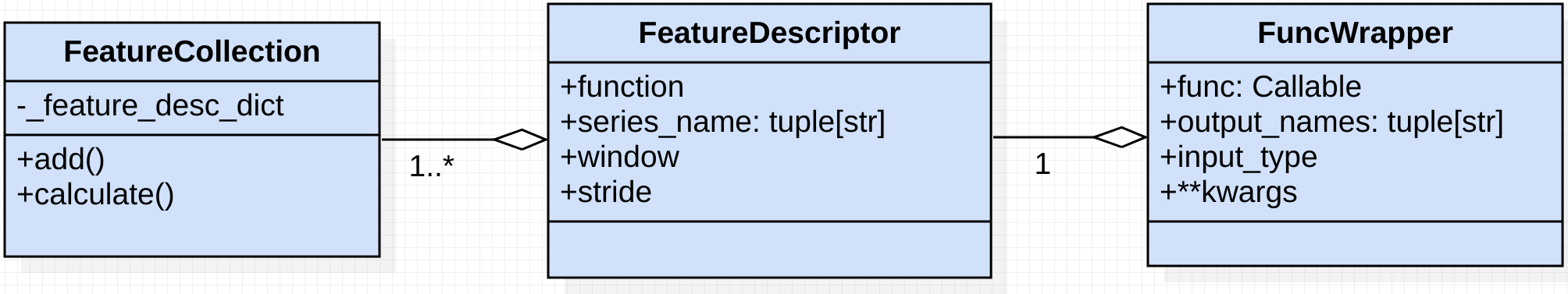}
   \caption{}
   \label{fig:features_uml}
\end{subfigure}

\caption[UML overview]{UML diagram of the (a) \texttt{tsflex.processing} and (b) \texttt{tsflex.features} submodule.}
\label{fig:uml}
\end{figure}

%Give a short overview of the overall software architecture; provide a pictorial component overview or similar (if possible). If necessary provide implementation details.
\texttt{tsflex} consists of two separated entities, i.e., a processing and a feature extraction submodule. The following subsections describe the architecture of both submodules, visually aided by figure~\ref{fig:uml}.

Remark that these two submodules work on a different scope. 
The processing submodule works on full sequences, i.e., full scope, whereas the feature extraction submodule works on strided windows, i.e., restricted scope.
%Moreover, processing operations are performed sequentially (and are thus not multithreaded) while feature functions can be executed in any order (and thus can also run in parallel).

\subsubsection{Processing submodule}
Figure~\ref{fig:processing_uml} depicts the main components of the\nolinebreak\hspace{\fill}\linebreak \texttt{tsflex.processing} submodule. The processing functionality is provided by the \texttt{SeriesPipeline} which contains one or multiple \texttt{SeriesProcessor} steps. The processing steps are applied sequentially on the data that is passed to the processing pipeline. This sequential order is crucial as the processing operations can create new series or update existing ones, which can be used in the succeeding steps, e.g., first applying a filter-processor and in the next steps decomposing that filtered signal. We summarize the objective of each component:

\begin{itemize}
    \item \texttt{SeriesPipeline}: serves as a pipeline, withholding the to-be-sequentially-applied \textit{processing steps}. 
    
    \item \texttt{SeriesProcessor}: an instance of this class describes a \textit{processing step}. 
    
    A processing step is defined by a \texttt{function} (the\nolinebreak\hspace{\fill}\linebreak \textit{Callable} processing-function), \texttt{series\_names} (the\nolinebreak\hspace{\fill}\linebreak \textit{name(s)} of the series that should be processed), and \texttt{**kwargs} (optional keyword arguments for \texttt{function}).
\end{itemize}

\subsubsection{Feature extraction submodule}
Figure~\ref{fig:features_uml} depicts the \texttt{tsflex.features} components. The feature extraction functionality is provided by a\nolinebreak\hspace{\fill}\linebreak \texttt{FeatureCollection} that contains one or multiple\nolinebreak\hspace{\fill}\linebreak \texttt{FeatureDescriptor}s. The features are calculated (possibly in parallel) on the data that is passed to the feature collection. We describe the objective of each component:

\begin{itemize}
    \item \texttt{FeatureCollection}: serves as a registry, withholding the to-be-calculated \textit{features}.
    
    \item \texttt{FeatureDescriptor}: an instance of this class describes a \textit{feature}.
    
    A feature is defined by a \texttt{series\_name} (the \textit{name(s)} of the input series on which the feature-function will operate), \texttt{function} (the \textit{Callable} feature-function), and \texttt{window} and \texttt{stride} (the sequence-index based window and stride range).
    
    \item \texttt{FuncWrapper}: a wrapper around \textit{Callable} functions, intended for advanced feature-function configuration (e.g., customizing feature output-names, passing\nolinebreak\hspace{\fill}\linebreak \texttt{**kwargs} to feature functions), and defining the function input data-type (i.e., \texttt{numpy.array} or\nolinebreak\hspace{\fill}\linebreak
    \texttt{pandas.Series}).
\end{itemize}

\subsection{Software Functionalities}
\label{subsec:software_functionalities}
%Present the major functionalities of the software.
In the sections below, we further detail the processing and feature extraction functionalities, together with other utilities of \texttt{tsflex}. 

% The two core functionalities of \texttt{tsflex} are time series \textbf{processing} and \textbf{feature extraction}.
% Both are decoupled from each other, i.e., there is no direct interplay between \texttt{tsflex.processing} and \texttt{tsflex.features}. However, they can be easily applied in a sequential manner.
% These two functionalities were implemented according to the core principles of \texttt{tsflex}, i.e, being \textbf{time series first} \& \textbf{flexible}.

\subsubsection{Processing}
The processing functionality is concerned with either transforming (i.e., replacing) sequences or creating new ones. \texttt{tsflex} provides flexible processing by accepting a generic \href{https://predict-idlab.github.io/tsflex/processing/index.html\#processing-functions}{processing function prototype}.
% \begin{verbnobox}[\small]
%   function(*series: pd.Series, **kwargs)
%      -> Union[np.ndarray, pd.Series, pd.DataFrame, List[pd.Series]]:
% \end{verbnobox}
Such processing functions should take one or multiple sequences as input, followed by optional keyword arguments. This generic processing function prototype enables compatibility with many existing libraries, e.g., \href{https://docs.scipy.org/doc/scipy/reference/reference/signal.html#module-scipy.signal}{\texttt{scipy.signal}}~\cite{2020SciPy-NMeth},\nolinebreak\hspace{\fill}\linebreak \href{https://www.statsmodels.org/stable/tsa.html#time-series-filters}{\texttt{statsmodels.tsa}}~\cite{seabold2010statsmodels}\footnote{Processing functions can return an arbitrary amount of sequences; \texttt{tsflex} supports one-to-one, one-to-many, many-to-one, and many-to-many functions; see \url{https://predict-idlab.github.io/tsflex/processing/index.html\#versatile-processing-functions}}.

\subsubsection{Feature extraction}
The feature extraction functionality is concerned with calculating features on strided-rolling windows. \texttt{tsflex} was designed to define the window and stride arguments in the same unit as the sequence-index its datatype (e.g., window=\textit{``5min''} and stride=\textit{``30s''} for time-indexed sequences, or window=\textit{300} and stride=\textit{30} for numeric-indexed sequences). As existing libraries define the window and stride in terms of number of samples~\cite{burns2018seglearn, christ2018tsfresh, barandas2020tsfel}, they implicitly assume that the sampling rate is fixed and there are no gaps. 
\texttt{tsflex}'s \textit{flexibility} is a direct consequence of not making such assumptions; by default, features can be extracted on multivariate time series with varying sampling rates and even gaps\footnote{It is the feature-function its responsibility to handle such cases correctly. 
Note that a feature-function can easily be made robust using the \texttt{make\_robust} wrapper from \texttt{tsflex.features.utils}.}. In addition, \texttt{tsflex} supports a wide range of feature functions, again enabling compatibility with many existing libraries, e.g., \href{https://numpy.org/doc/stable/reference/routines.html}{\texttt{numpy}},  \href{https://docs.scipy.org/doc/scipy/reference/tutorial/stats.html}{\texttt{scipy.stats}}, \href{https://github.com/blue-yonder/tsfresh/blob/main/tsfresh/feature_extraction/feature_calculators.py}{\texttt{tsfresh}}\footnote{For more details on how \texttt{tsflex} integrates with existing libraries, consult \href{https://github.com/predict-idlab/tsflex/tree/main/examples}{our integration notebooks}.\label{note:integration_notebooks}}, 
\href{https://dmbee.github.io/seglearn/feature_functions.html}{\texttt{seglearn}}\textsuperscript{\ref{note:integration_notebooks}}, \href{https://tsfel.readthedocs.io/en/latest/descriptions/modules/tsfel.feature_extraction.html#module-tsfel.feature_extraction.features}{\texttt{tsfel}}\textsuperscript{\ref{note:integration_notebooks}}.

\subsubsection{Other functionalities}
\texttt{tsflex} serves various additional functionalities, such as embedded serialization, execution time logging, native support for categorical \& time based data, and handling of time series data in chunks. Chunking of sequence data can be performed by calling the \texttt{chunk\_data} function from the \href{https://predict-idlab.github.io/tsflex/chunking/index.html}{\texttt{tsflex.chunking}} submodule. Processing \& extracting features on chunked data produces lower memory peaks, enabling time series handling in constrained environments (e.g., streaming, edge devices~\cite{shi2016promise}). Additionally, chunking allows parallelizing the sequential processing. Lastly, the \texttt{FeatureCollection} its \href{https://predict-idlab.github.io/tsflex/features/index.html#tsflex.features.FeatureCollection.reduce}{\texttt{reduce(feat\_cols\_to\_keep)}} method returns a new \texttt{FeatureCollection} instance, withholding the subset of features that constitute the output column names listed in \texttt{feat\_cols\_to\_keep}.

\paragraph{Unit tests} The provided functionalities of \texttt{tsflex} are extensively tested through unit testing. For example, these tests assure that the functions should perform view-based operations, that \texttt{tsflex} handles categorical and time-based data, and that feature-functions are not allowed to change the view-based input data. Every claim about \texttt{tsflex} we make in this paper is backed by unit testing.

\subsection{Limitations}
% OLD: The little assumptions that \texttt{tsflex} makes, pose some (albeit logical) limitations on the supported data format. These limitations are (1) each series must have a \texttt{pd.DatetimeIndex} \textbf{that increases monotonically}, and (2) \textbf{no duplicate series names} are allowed. For both limitations, there are simple remediations; one can (1) apply \href{https://pandas.pydata.org/pandas-docs/stable/reference/api/pandas.DataFrame.sort_index.html}{\texttt{sort\_index()}} on the not-monotonically increasing time-indexed data, and (2) rename the series.
% A potential third limitation might be that \texttt{tsflex} is intended to work on \textit{flat single-index data} such as a list of series or a wide-dataframe. Hence, there is no native support for a long-dataframe. A convenient solution for this is to transform the long-dataframe into a list of series.\footnote{Note that this conversion will cause a significant memory peak. More information and code for this conversion can be found \href{https://predict-idlab.github.io/tsflex/\#wide-vs-long-data}{in our documentation}.}
Currently, there is no agreed standard for time series in Python~\cite{max_time_series}. The main cause for this disagreement is that each format has its own benefits and disadvantages. An in-depth discussion about this topic is out of scope for this paper. For \texttt{tsflex}, we made the design decision to operate on single-indexed wide/flat data (such as a list of series or a wide-dataframe) whose index represents the sequence-position. In our opinion, this data format is most intuitive to wrangle with, e.g., slicing, visualizing, processing. Therefore, two limitations of \texttt{tsflex} are that it (1) does not support long data, nor (2) multi-indexed data (and columns). Remark that a long-dataframe can be transformed into a list of series (that has the same in-memory size). %More info on data formats can be found in our \href{https://predict-idlab.github.io/tsflex/\#data-formats}{documentation}\footnote{See also \url{https://predict-idlab.github.io/tsflex/\#data-formats} and \url{https://predict-idlab.github.io/tsflex/features/index.html\#limitations}}. 
A third limitation is that \texttt{tsflex} uses sequence-names as identifiers, resulting in the assumption that each sequence should have a unique name.

%Additionally, tsflex does not support . A final limitation is that each sequence must have a unique name.

% \begin{enumerate}
%     \item \textbf{Data structure assumptions:}
%     \begin{enumerate}
%         \item No support for multi-index data (or multi-index columns)
%         \item Each series must have a unique name (and a sequence index)
%         \item Assumes flat data as input (where the index represents the sequence position)
%     \end{enumerate}
%     % \item featurediscriptor cannot return a variable amount of features (based on input length)
% \end{enumerate}

% \texttt{tsflex} is \textbf{flexible}, this is reflected by (1) the versatile input that the processing and feature functions support, (2) the  little assumptions it makes about the time-series data, and (3) the wide range of additional supported functions (e.g., logging of execution times, serialization).  

\section{Illustrative Examples}
\label{sec3:examples}
% Provide at least one illustrative example to demonstrate the major functions.

As illustrative example, we provide three snippets containing working code~\footnote{Online version: \url{https://github.com/predict-idlab/tsflex/blob/main/examples/tsflex_paper.ipynb}.}.

\paragraph{Data loading}
Listing~\ref{listing:data_loading} fetches the data for the examples. In total, three \texttt{pd.DataFrame}s are loaded, containing multimodal data of different sampling rates. This data is an excerpt of a wrist-worn wearable from the WESAD study~\cite{schmidt2018introducing}. The characteristics of the dataframes are summarized in Table~\ref{tab:example-data-info}. 
The \texttt{df\_tmp} dataframe withholds skin temperature data, 
%\texttt{df\_gsr} contains ElectroDermal Activity (EDA) or Galvanic Skin Response (GSR) data, indicating the skin its conductance level; 
\texttt{df\_acc} withholds accelerometer data along the 3 movement axes, and \texttt{df\_ibi} contains the Inter-Beat-Interval (IBI) data, representing the time between two consecutive heartbeats. Remark that IBI data is only available when two consecutive, successfully detected beats took place, making IBI an irregularly sampled series.

\begin{listing}[ht]
\begin{lstlisting}[language=Python]
from tsflex.utils.data import load_empatica_data
df_tmp, df_acc, df_ibi = load_empatica_data(["tmp","acc","ibi"])
\end{lstlisting}
\caption{Data loading code.}
\label{listing:data_loading}
\end{listing}

\begin{table}[ht]
\centering
\footnotesize
\begin{tabular}{lp{22mm}lp{21mm}}
\cline{2-4}
\textbf{} & \textbf{columns} & \textbf{shape} & \textbf{sampling rate} \\ \hline
\textbf{df\_tmp} & {[}TMP{]} & (30200, 1) & 4.0 Hz \\
% \textbf{df\_gsr} & {[}EDA{]} & (30204, 1) & 4.0 Hz \\
\textbf{df\_acc} & {[}ACC\_x, ACC\_y, ACC\_z{]} & (241620, 3) & 32.0 Hz \\
\textbf{df\_ibi} & {[}IBI{]} & (1230, 1) & Irregularly sampled \\ \hline
\end{tabular}
\caption{Properties of the data used in the examples.}
\label{tab:example-data-info}
\end{table}

\subsection{Processing}
Listing~\ref{listing:processing_example} shows how various processing steps are applied on the loaded data. For each processing step a\nolinebreak\hspace{\fill}\linebreak \texttt{SeriesProcessor} object is created, which records the series names\footnote{When a processing function should be applied on multiple series, a list should be passed to the \texttt{series\_names} argument. When a processing function handles multiple series as input, a tuple (or a list thereof) should be passed to the \texttt{series\_names} argument.} (i.e., the names of the sequences that should be processed) and the optional keyword arguments. Observe that the \texttt{smv} function creates a new series. 
%When calling the \texttt{process} method of the constructed processing pipeline, the original accelerometer columns will be dropped due to the \texttt{drop\_keys} argument. Hence, the output of this example is a list of series containing the following three processed series; \texttt{TMP}, \texttt{IBI}, and \texttt{ACC\_SMV}.

\begin{listing}%[ht]
\begin{lstlisting}[language=Python]
import pandas as pd; import numpy as np 
from scipy.signal import savgol_filter
from tsflex.processing import SeriesProcessor, SeriesPipeline

# Create the processing functions
def clip_data(sig: pd.Series, min_val=None, max_val=None):
    return np.clip(sig,  a_min=min_val, a_max=max_val)

def smv(*sigs) -> pd.Series:
    sig_prefixes = set(s.name.split('_')[0] for s in sigs)
    res = np.sqrt(np.sum([np.square(s) for s in sigs], axis=0))
    return pd.Series(res, index=sigs[0].index, 
        name='|'.join(sig_prefixes)+'_'+'SMV')

# Create the series processors (with their keyword arguments)
tmp_clippper = SeriesProcessor(clip_data, series_names="TMP", 
    max_val=35)
acc_savgol = SeriesProcessor(
    savgol_filter, ["ACC_x", "ACC_y", "ACC_z"], window_length=33, 
    polyorder=2
)
acc_smv = SeriesProcessor(smv, ("ACC_x", "ACC_y", "ACC_z"))

# Create the series pipeline & process the data
series_pipe = SeriesPipeline([tmp_clippper, acc_savgol, acc_smv])
out_data = series_pipe.process([df_acc, df_tmp, df_ibi])
\end{lstlisting}
\caption{Processing example. Continuation of code snippet~\ref{listing:data_loading}.}
\label{listing:processing_example}
\end{listing}

\subsection{Feature extraction}
Listing~\ref{listing:feature_extraction_example} shows how feature extraction can be performed on the previously processed data. Two\nolinebreak\hspace{\fill}\linebreak \texttt{MultipleFeatureDescriptors}\footnote{\texttt{MultipleFeatureDescriptors} are a convenient way to define features containing multiple functions, series names, windows, and strides.} are created; the first defines some general statistical and spectral features on the \texttt{ACC\_SMV} and \texttt{TMP} signal for two different windows, and the second defines a robust version\footnote{It is important to make the feature functions robust as there may be empty windows for the \texttt{IBI} data. In such cases, the \texttt{make\_robust} wrapper avoids that an error is thrown and returns \texttt{NaN} instead.} of some  statistical features (and the number of samples) for the \texttt{IBI} signal. Remark that in \texttt{general\_feats}, \texttt{seglearn} feature-functions are imported and wrapped in a convenient manner.
These two descriptor objects are enclosed in a feature collection, which is used for extracting (i.e., calculating) the features. The \texttt{approve\_sparsity} flag enables the user to explicitly acknowledge that there might be sparse data, i.e., irregularly sampled data. 
Setting this flag avoids warnings that are raised in case of sparsity. 

\begin{listing}%[ht]
\begin{lstlisting}[language=Python]
from tsflex.features import MultipleFeatureDescriptors, 
    FeatureCollection
from tsflex.features.integrations 
    import seglearn_feature_dict_wrapper
from tsflex.features.utils import make_robust

# Import / create the feature functions
from seglearn.feature_functions import base_features
def area(sig: np.ndarray): return np.sum(np.abs(sig))

# Create the feature descriptors
general_feats = MultipleFeatureDescriptors(
        functions=seglearn_feature_dict_wrapper(base_features()) 
            + [area],
        series_names=["ACC_SMV", "TMP"],
        windows=["5min", "2.5min"], strides="2min",
)
ibi_feats = MultipleFeatureDescriptors(
    [make_robust(f) for f in [np.min, np.max, np.mean, np.std]] +
        [len],
    series_names="IBI", windows="5min", strides="2min"
)

# Create the feature collection & calculate the features
fc = FeatureCollection([general_feats, ibi_feats])
feat_df = fc.calculate(out_data, return_df=True, 
    approve_sparsity=True)
\end{lstlisting}
\caption{Feature extraction example. Continuation of code snippet~\ref{listing:processing_example}.}
\label{listing:feature_extraction_example}
\end{listing}

% Optional: you may include one explanatory video that will appear next to your article, in the right hand side panel. (Please upload any video as a single supplementary file with your article. Only one MP4 formatted, with 50MB maximum size, video is possible per article. Recommended video dimensions are 640 x 480 at a maximum of 30 frames/second. Prior to submission please test and validate your .mp4 file at $ http://elsevier-apps.sciverse.com/GadgetVideoPodcastPlayerWeb/verification$. This tool will display your video exactly in the same way as it will appear on ScienceDirect.).

\section{Impact}
\label{sec4:impact}
% \textbf{This is the main section of the article and the reviewers weight the description here appropriately}
% Indicate in what way new research questions can be pursued as a result of the software (if any).
% Indicate in what way, and to what extent, the pursuit of existing research questions is improved (if so).
% Indicate in what way the software has changed the daily practice of its users (if so).
% Indicate how widespread the use of the software is within and outside the intended user group.

% Indicate in what way the software is used in commercial settings and/or how it led to the creation of spin-off companies (if so).

% \texttt{tsflex} arose as current time series feature extraction toolkits make too strong assumptions, and were rather inefficient in both computation time and memory usage. The flexibility issue is detailed in subsection~\ref{sec:impact-functionlaities}. 

We first indicate the impact of \texttt{tsflex} by positioning it among other packages. Then, we present \texttt{tsflex}'s performance in terms of memory usage and computation time, and compare these results to related packages. We conclude with some examples and references to notebooks which highlight the cross-domain applicability of \texttt{tsflex} for time series. 

\subsection{Functionalities}\label{sec:impact-functionlaities}
Irregularly sampled data is ubiquitous. However, most existing time series toolkits assume that either the user segments the data in valid chunks or that the data is regularly sampled. The former induces a significant user burden, whilst the latter is a fairly strong assumption. By employing a \texttt{sequence range based} window-stride approach and thus not a \textit{sample based} one, \texttt{tsflex} interoperates natively with irregularly sampled sequence data.
We position such functionalities of \texttt{tsflex} against other related packages in Table~\ref{tab:comparison}. Remark that \texttt{tsflex} is the only package that (1) allows defining multiple window-stride combinations, (2) can operate on non-numerical data, and (3) serves time-based chunking functionalities. 
Moreover, except for \texttt{tsfresh}, \texttt{tsflex} is the only other library that maintains the index of the data, encouraging index based analysis of the obtained outputs.
We refer to \href{https://github.com/predict-idlab/tsflex/tree/main/examples}{example notebooks} for more concrete illustrations of these functionalities.%~\footnote{\url{https://github.com/predict-idlab/tsflex/tree/main/examples}}.

\begin{table*}[ht]
\footnotesize
\centering
% \resizebox{\textwidth}{!}{%
% \renewcommand{\arraystretch}{1.15}
\begin{tabular}{@{}lP{0.12\textwidth}P{0.12\textwidth}P{0.12\textwidth}P{0.12\textwidth}P{0.12\textwidth}@{}}
\cmidrule(l){2-6}
\textbf{Properties} & \textbf{tsflex} & \textbf{seglearn} & \textbf{tsfresh} & \textbf{TSFEL} & \textbf{kats} \\ \midrule
\textit{\textbf{General}} & \multicolumn{1}{l}{} & \multicolumn{1}{l}{} & \multicolumn{1}{l}{} & \multicolumn{1}{l}{} & \multicolumn{1}{l}{} \\
Time column requirements & Any - sortable & Any - sorted & Any - sortable & Any - sorted & \href{https://pandas.pydata.org/docs/reference/api/pandas.DatetimeIndex.html#pandas-datetimeindex}{Datetime index} \\
Multivariate time series & \cmark & \cmark & \cmark & \cmark & \cmark \\
Unevenly sampled data & \cmark  & \xmark & \xmark & \xmark  & \cmark \\
Time column maintenance & \cmark & \xmark & \cmark & \xmark & \xmark  \\
Retain output names & \cmark & \cmark & \cmark & \cmark & \xmark \\
Multiprocessing & \cmark & \xmark & \cmark & \cmark & \xmark  \\
Operation execution time logging & \cmark  & \xmark & \xmark & \xmark & \xmark \\
Chunking (multiple) time series & \cmark & \xmark & \xmark & \xmark & \xmark  \\[0.5cm]

\textit{\textbf{Feature extraction}} & \textit{} & \textit{} & \textit{} & \textit{} & \textit{} \\ 
Strided-window definition format & Sequence index range & Sample-based & Sample-based & Sample-based & Na. \\
Strided-window feature extraction & \cmark & \cmark & \cmark & \cmark & \xmark \\
Multiple stride-window combinations & \cmark & \xmark & \xmark & \xmark & \xmark \\
Custom features & \cmark & \cmark & \cmark & \cmark & \xmark \\
One-to-one functions & \cmark & \cmark & \cmark & \cmark & \cmark \\
One-to-many functions & \cmark & \cmark & \cmark & \cmark & \cmark \\
Many-to-one functions & \cmark & \cmark & \xmark & \xmark & \xmark  \\
Many-to-many functions & \cmark & \xmark & \xmark & \xmark & \xmark \\
% Support streaming & \cmark & \cmark & \xmark & \cmark  & \xmark \\
Categorical data & \cmark  & \xmark & \xmark  & \xmark & \xmark \\
Datatype preservation & \cmark  & \xmark & \xmark  & \xmark & \xmark \\ \bottomrule
\end{tabular}
% }
\caption{Comparison of \texttt{tsflex} against other relevant packages. The \textit{``$X$-to-$Y$ functions''} in the Properties column with $X, Y \in \{one, many\}$ represent the feature input-to-output relationship; $X$=``one" denotes single-series input, whereas $X$=``many'' represents multivariate inputs. When $Y$=``one'' a single feature is returned, whilst the $Y$=``many'' returns multiple features. More info about these versatile functions can be found \protect\href{https://predict-idlab.github.io/tsflex/features/index.html\#versatile-functions}{here}. An online version of this table is shown \protect\href{https://predict-idlab.github.io/tsflex/\#comparison}{here}.}
\label{tab:comparison}
\end{table*}

\subsection{Feature extraction performance}\label{sec:feat_extr_performance}
Considering all Python toolkits, eligible for strided-rolling feature extraction~\cite{burns2018seglearn, christ2018tsfresh, barandas2020tsfel}, only \texttt{seglearn} mentions toolkit-performance by comparing their computation time and model accuracy with other packages. However, for real-world applicability, computational efficiency is of utmost importance. Therefore, we benchmarked \texttt{tsflex} its memory usage and runtime against other libraries and open-sourced the benchmarking codebase at \href{https://github.com/predict-idlab/tsflex-benchmarking}{this repository}\footnote{We decided to only benchmark feature extraction, as this is the most advanced functionality of \texttt{tsflex}. In our experience, the processing functionality is rather straightforward and thus more dependent on the processing functions. However, empirical results indicate that we have a significant efficiency advantage over other existing packages when parallel processing is performed on chunked data.} to encourage effortless benchmarking of \texttt{tsflex} on other use cases (e.g., edge devices, extremely large datasets, streaming use cases).

Profiling is realized by using the \verb|VizTracer|~\cite{viztracer} package with the \texttt{\href{https://github.com/gaogaotiantian/vizplugins}{VizPlugins}} add-on. The benchmark dataset is a synthetically generated dataframe consisting of 5 channels and spans 1 hour. Its values have the numerical \href{https://numpy.org/doc/1.20/reference/arrays.scalars.html\#numpy.float32}{\texttt{numpy.float32}} data type. To comply with the assumptions that other toolkits make, each modality is sampled at 1000Hz and does not contain gaps. The toolkit are configured to extract the same features using a window-stride of 30s-10s, respectively. The benchmark process follows these steps for each toolkit-feature-extraction configuration:
\begin{enumerate}
  \item Each toolkit feature extraction script is called 20 times to average out the memory usage and runtime\footnote{Remark that by recalling the script in separate runs, no caching or memory is shared among executions.}.
  \item Script execution:
  \begin{enumerate}
      \item Construct the synthetic \texttt{pd.DataFrame} benchmark data
      \item \texttt{VizTracer} starts logging
      \item Create the feature extraction configuration
      \item Extract and store the features
      \item \texttt{VizTracer} stops logging
      \item Write the \texttt{VizTracer} profile-results to a JSON-file
  \end{enumerate}
\end{enumerate}

The profile JSONs were collected on a server with an \textit{Intel Xeon E5-2650 v2 @ 2.60GHz} CPU and {SAMSUNG M393B1G73QH0-CMA DDR3 1600MT/s} RAM, with\nolinebreak\hspace{\fill}\linebreak \textit{Ubuntu 18.04.5 LTS x86\_64} as operating system. Other running processes were limited to a minimum.

\begin{figure*}[ht]
\centering
\includegraphics[width=0.85\textwidth]{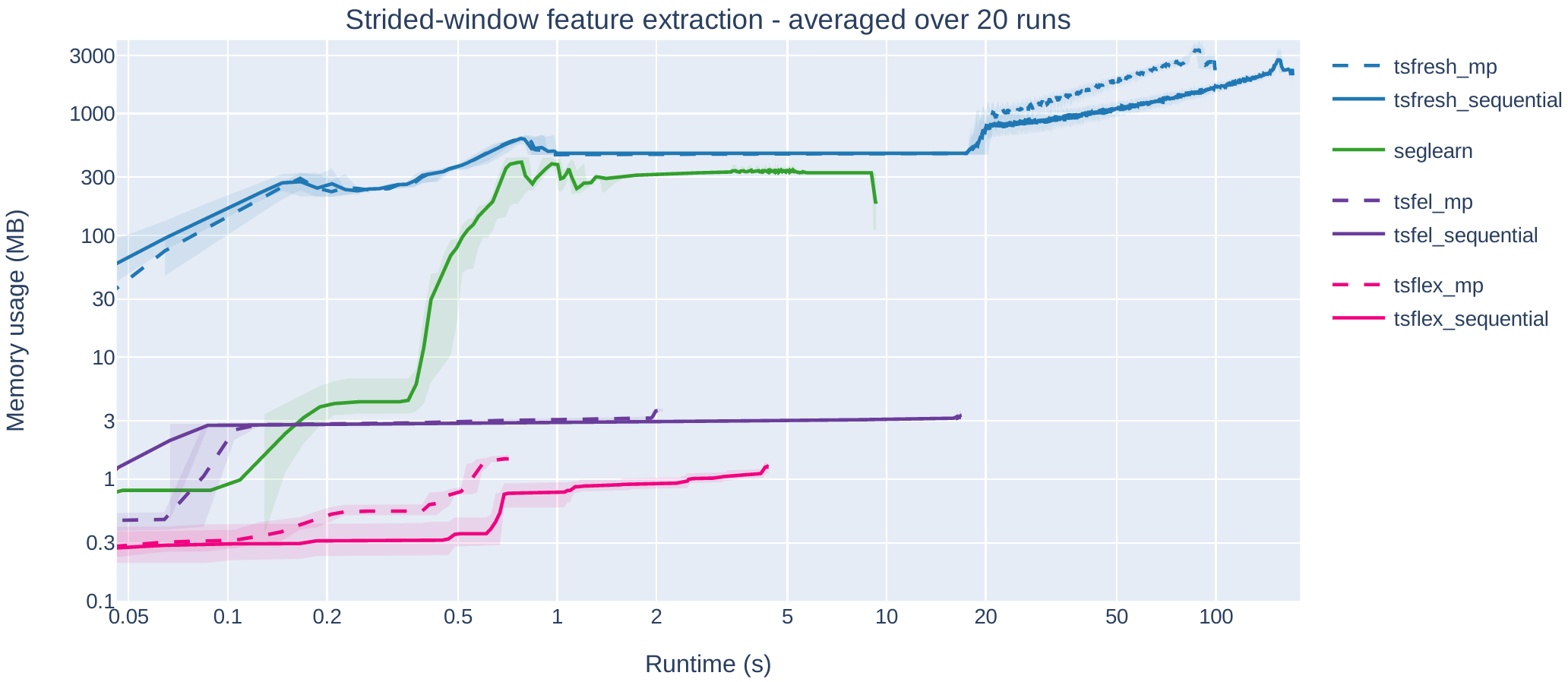}
\caption[Average memory usage over time for a feature extraction problem with a fixed window and stride]{Average memory usage over time for a feature extraction task on evenly sampled data with a fixed window and stride. The origin of the runtime and memory usage axis starts directly after the synthetic data was constructed; the feature extraction configuration is then initialized and called on the data. As noted in Table~\ref{tab:comparison}, only \texttt{seglearn v1.2.3} \protect\cite{burns2018seglearn}, \texttt{tsfresh v0.18.0}~\cite{christ2018tsfresh}, and \texttt{TSFEL v0.1.4}~\cite{barandas2020tsfel} support defining a (sample-based) window and stride, making this comparison fair as the data for this benchmark is evenly sampled. For reference, the allocated memory for the data was $96.4MB$. An interactive version (where you can switch to linear axes) of this figure is shown \protect\href{https://predict-idlab.github.io/tsflex/\#benchmark}{here}.}
\label{fig:benchmark}
\end{figure*}

Figure~\ref{fig:benchmark} depicts the aggregated JSON-file results and table~\ref{tab:benchmark_results} summarizes the main outcomes of this visualization. For this use-case, \texttt{tsflex} is $\sim3\times$ faster than its closest competitor in both the sequential and multiprocessing variant. The peak memory usage is of particular interest, as this determines the minimum amount of RAM a system should have. \texttt{tsflex} and \texttt{TSFEL} apply view-based operations on the data, making them significantly more memory efficient than other packages. Here again, \texttt{tsflex} requires $\sim 2.5 \times$ less memory than \texttt{TSFEL}. Note that \texttt{tsfresh} first expands the data into a tsfresh-compatible format before applying feature extraction. This results in a slope in the logarithmic domain from second 15 to second 80-150.

\begin{table}[ht]
\resizebox{0.48\textwidth}{!}{%
\begin{tabular}{@{}rcccc@{}}
\cline{2-5}
 & \textbf{tsflex} & \textbf{TSFEL} & \textbf{seglearn} & \textbf{tsfresh} \\ \midrule
\textbf{\begin{tabular}[c]{@{}r@{}}mean peak memory \\ usage (MB $\pm$ std)\end{tabular}} & \multicolumn{1}{l}{} & \multicolumn{1}{l}{} & \multicolumn{1}{l}{} & \multicolumn{1}{l}{} \\
\multicolumn{1}{r}{sequential} & \textbf{1.3 $\pm$ 0.1} & 3.5 $\pm$ 0.3  & 435.3 $\pm$ 1.5 & 3540 $\pm$ 13.9 \\
\multicolumn{1}{r}{multiprocessing} & \textbf{1.5 $\pm$ 0.1} & 3.7 $\pm$ 0.1  & / & 4044 $\pm$ 14.4 \\[0.5cm]
%  &  &  &  &  \\
\textbf{\begin{tabular}[c]{@{}r@{}}mean runtime \\ (s $\pm$ std)\end{tabular}} & \multicolumn{1}{l}{} & \multicolumn{1}{l}{} & \multicolumn{1}{l}{} & \multicolumn{1}{l}{} \\\multicolumn{1}{r}{sequential} & \textbf{4.3 $\pm$ 0.1}  & 16.4 $\pm$ 0.8  & 9.2 $\pm$ 0.1 & 169.8 $\pm$ 1.6 \\
\multicolumn{1}{r}{multiprocessing} & \textbf{0.7 $\pm$ 0.0}  & 2.1 $\pm$ 0.0 & / & 98.5 $\pm$ 1.2 \\ \bottomrule
\end{tabular}%
}
\caption{Tabular summary of \texttt{VizTracer} benchmarks, depicted in Figure \ref{fig:benchmark}.}
\label{tab:benchmark_results}
\end{table}

\subsection{Applicability}
\texttt{tsflex} is a domain-independent package, enabling\nolinebreak\hspace{\fill}\linebreak broad applicability\footnote{The cross-domain applicability is highlighted by the examples: \url{https://github.com/predict-idlab/tsflex/tree/main/examples}}. For example, this package is already used in multiple projects such as wearable-based stress monitoring, automatic sleep staging, occupancy detection in buildings, and anomaly detection. \texttt{tsflex}'s computational efficiency (in both execution time and memory usage) also paves the way towards applicability in constrained environments, such as streaming or edge computing~\cite{shi2016promise}.

\section{Conclusions}
\label{sec5:conclusion}
Time series processing and feature extraction are arguably the most important steps in classical machine learning pipelines. However, existing packages are limited in their applicability as they make strong assumptions about the underlying data types and data structure. %This adds unnecessary overhead when using them on as specific use case. %This makes it inconvenient to use those packages on different datasets.
% Moreover, these packages . 
Furthermore, these toolkits do not prioritize memory and runtime efficiency, creating unnecessary overheads. These existing packages also tend to focus on including numerous feature functions instead of conveniently integrating with other libraries.
% However, current packages are limited as they lack the flexibility to be used on different datasets. For instance, these packages make strong assumptions about the underlying data types and structure of the data. 
We argue that there is a need for a more permissive toolkit, which concentrates on the essentials.
%
%We would argue that those package focus more on adding feature functions (ML models, integrated feature, features selection, etc..) and there is the need for a package that focuses on the essence of feature creation and data preprocessing. By keeping the scope limited and focusion on doing the things right... 
%
Therefore, we present \texttt{tsflex}, a Python package that focuses on processing and feature extraction for time series. 
% Both are crucial steps in machine learning pipelines. 
%To enable broad applicability, \texttt{tsflex} tries to be as permissive as possible with the data format window-stride options. 
% To the best of our knowledge, \texttt{tsflex} is the first toolkit using sequence-index based parameters for feature extraction and data chunking, resulting in an increased flexibility and more user convenience
%\texttt{tsflex} focuses on getting these fundamentals right. 
This paper describes the functionalities and performance of \texttt{tsflex} and compares it to other packages. We show that \texttt{tsflex} is more permissive than existing Python toolkits, and benchmarking indicates it is over $50\%$ more efficient than comparable work in both runtime and memory usage. The increased flexibility is realized by leveraging sequence-index based arguments and is reflected in the few assumptions that this library makes. We believe that \texttt{tsflex}'s integration with other libraries, together with its advanced functionalities, e.g., chunking, comprehensible feature output names, enables real-world, cross-domain applicability.

\section{Conflict of Interest}
%Please select the appropriate text:

%Potential conflict of interest exists:
%We wish to draw the attention of the Editor to the following facts, which may be considered as potential conflicts of interest, and to significant financial contributions to this work. The nature of potential conflict of interest is described below: [Describe conflict of interest]

%No conflict of interest exists:
%We wish to confirm that there are no known conflicts of interest associated with this publication and there has been no significant financial support for this work that could have influenced its outcome.

The authors declare that they have no conflict of interest.

\section*{Acknowledgements}
\label{acknowledgments}
% OLD: Emiel Deprost is funded by a doctoral fellowship of the Research Foundation – Flanders (FWO). Jonas Van Der Donckt is funded by a doctoral fellowship of Ghent University (BOF). 
Jonas Van Der Donckt and Emiel Deprost are funded by a doctoral fellowship of the Research Foundation – Flanders (FWO). Part of this work is done in the scope of the imec.ICON COSMO (HBC.2018.0531), imec.AAA Context-aware health monitoring, and VLAIO PoC Nervocity.



%% References:
\bibliographystyle{elsarticle-num} 
\bibliography{references}

% TODO:
% Please add the reference to the software repository if DOI for software  is available. 

% \section*{Current executable software version}
% \label{seq:curr_software_version}

% %Ancillary data table required for sub version of the executable software: (x.1, x.2 etc.) kindly replace examples in right column with the correct information about your executables, and leave the left column as it is.

% \begin{table}[!h]
% \begin{tabular}{|l|p{6.5cm}|p{6.5cm}|}
% \hline
% \textbf{Nr.} & \textbf{(Executable) software metadata description} & \textbf{Please fill in this column} \\
% \hline
% S1 & Current software version & v0.2.3 \\
% \hline
% S2 & Permanent link to executables of this version  & \url{https://github.com/predict-idlab/tsflex} \\
% \hline
% S3 & Legal Software License & MIT \\
% \hline
% S4 & Computing platforms/Operating Systems & Linux, OS X, Microsoft Windows \\
% \hline
% S5 & Installation requirements \& dependencies & Python3.7+ \\
% \hline
% S6 & If available, link to user manual - if formally published include a reference to the publication in the reference list & \url{https://predict-idlab.github.io/tsflex} \\
% \hline
% S7 & Support email for questions & jonvdrdo.vanderdonckt@ugent.be \\
% \hline
% \end{tabular}
% \caption{Software metadata (optional)}
% \label{software_metadata} 
% \end{table}

\end{document}